\documentclass{ieeeaccess}
\usepackage{cite}
\usepackage{amsmath,amssymb,amsfonts}
\usepackage{algorithmic}
\usepackage{graphicx}
\usepackage{textcomp}
\usepackage{cite}
\usepackage{amsmath,amssymb,amsfonts}
\usepackage{algorithmic}
\usepackage{graphicx}
\usepackage{textcomp}
\usepackage{xcolor}
\usepackage{multirow}
\usepackage{array,multirow}
\usepackage{booktabs}
\usepackage{graphicx}
\usepackage{subcaption}
\usepackage{bm}

\makeatletter
\AtBeginDocument{\DeclareMathVersion{bold}
\SetSymbolFont{operators}{bold}{T1}{times}{b}{n}
\SetSymbolFont{NewLetters}{bold}{T1}{times}{b}{it}
\SetMathAlphabet{\mathrm}{bold}{T1}{times}{b}{n}
\SetMathAlphabet{\mathit}{bold}{T1}{times}{b}{it}
\SetMathAlphabet{\mathbf}{bold}{T1}{times}{b}{n}
\SetMathAlphabet{\mathtt}{bold}{OT1}{pcr}{b}{n}
\SetSymbolFont{symbols}{bold}{OMS}{cmsy}{b}{n}
\renewcommand\boldmath{\@nomath\boldmath\mathversion{bold}}}
\makeatother

\def\BibTeX{{\rm B\kern-.05em{\sc i\kern-.025em b}\kern-.08em
    T\kern-.1667em\lower.7ex\hbox{E}\kern-.125emX}}

\begin{document}
\history{\phantom{Date of publication xxxx 00, 0000, date of current version xxxx 00, 0000.}}
\doi{\phantom{10.1109/ACCESS.2024.0429000}}

\title{Three Forensic Cues for JPEG AI Images}
\author{\uppercase{Sandra Bergmann}\authorrefmark{1},
\uppercase{Fabian Brand}\authorrefmark{2}, and \uppercase{Christian Riess}\authorrefmark{1}}

\address[1]{IT Security Infrastructures Lab, Friedrich-Alexander-Universität Erlangen-Nürnberg, Germany}
\address[2]{Multimedia Communications and Signal Processing Lab, Friedrich-Alexander-Universität Erlangen-Nürnberg, Germany}

\markboth
{S. Bergmann \headeretal: Three Forensic Cues for JPEG AI Images}
{S. Bergmann \headeretal: Three Forensic Cues for JPEG AI Images}

\corresp{Corresponding author: Sandra Bergmann (e-mail: sandra.daniela.bergmann@fau.de).}

\begin{abstract}
The JPEG standard was vastly successful. Currently, the first AI-based compression method ``JPEG AI'' will be standardized.
JPEG AI brings remarkable benefits. JPEG AI images exhibit impressive image quality at bitrates that are an order of
magnitude lower than images compressed with traditional JPEG.
However, forensic analysis of JPEG AI has to be completely re-thought:
forensic tools for traditional JPEG do not transfer to JPEG AI, and artifacts
from JPEG AI are easily confused with artifacts from artificially generated
images (``DeepFakes''). This creates a need for novel forensic approaches to
detection and distinction of JPEG AI images.

In this work, we make a first step towards a forensic JPEG AI toolset.
We propose three cues for forensic algorithms for JPEG AI. These
algorithms address three forensic questions:
first, we show that the JPEG AI preprocessing introduces correlations in the
color channels that do not occur in uncompressed images.
Second, we show that repeated compression of JPEG AI images leads to
diminishing distortion differences. This can be used to detect
recompression, in a spirit similar to some classic JPEG forensics methods.
Third, we show that the quantization of JPEG AI images in the latent space can
be used to distinguish real images with JPEG AI compression from synthetically
generated images.
The proposed methods are interpretable for a forensic analyst, and we hope that they
inspire further research in the forensics of AI-compressed images.
\end{abstract}


\titlepgskip=-21pt

\maketitle

\section{Introduction}
\label{sec:intro}
Lossy image compression is particularly efficient in storing digital images:
when an image is compressed in a lossy format, then only perceptually relevant
elements of an image are stored, and everything else is discarded in order to
save space. This process of transforming and reducing data is a rich trove of
information for the forensic analysis of images: compression traces can provide
cues about the origin of an image~\cite{Zhigang_2003, Shruti_2020_RoundingJPEG}, global image
transformations like recompression~\cite{Lukas_03_DJPEG,Popescu_Farid_05},
and local image
editing~\cite{CATNET, Verma_2023}. It is not by chance that most of the works on
compression forensics study traces from the JPEG algorithm, since JPEG is the
most widely used compression format. However, this may change in the future.
Recently, several AI-based image compression methods where proposed,
which achieve significantly lower bitrates at superior perceived image
quality~\cite{Balle_Var_Autoencoder, mbt2018, ms2020, mentzer2020high}.
As a consequence, the JPEG committee recently standardized an AI-based
compressor, called JPEG AI~\cite{JPEGAI_Ascenso, Alshina2024}.
JPEG AI is a general purpose codec, designed for use in diverse application
fields, e.g. visual surveillance, image collection storage, or media
distribution. JPEG AI is published as an international standard in
February 2025~\cite{Alshina2024}.

If JPEG AI finds widespread adoption, then this will also have great impact on
compression forensics. 
AI-based compression works differently than classical JPEG compression, and
hence forensics tools for traditional JPEG are not applicable to AI-based
compressors~\cite{Berthet_ICIP_AI_Compr}.
Moreover, JPEG AI leave visual artifacts, which may even change the semantic image
content~\cite{Hofer2024, Tsereh2024}.
To further complicate matters, a detector for generated images may confuse traces from JPEG AI in real images with traces from synthetically generated images~\cite{Cannas2024}. 
Hence, in order to brace for a presumably wide adoption of the new JPEG AI
standard, it will also be necessary to develop a new set of forensic tools for
JPEG AI images.

In this work, we make a step towards this goal.
We propose three cues for statistical forensics on JPEG AI images.
The proposed forensic traces address the detection of JPEG AI compression, the detection of JPEG AI recompression, and the discrimination of JPEG AI and AI-generated images.
More specifically, we propose:

\begin{enumerate}
\item \textbf{Color correlations for detection of compression.} JPEG AI compresses images in YUV color space, which slightly correlates the RGB color channels.
While a similar effect can be observed for traditional JPEG images, it nevertheless enables detection of JPEG AI compression particularly for stronger compression. 
\item \textbf{Rate-distortion cues for detecting recompression.} The
peak-signal-to-noise ratio (PSNR) of an image decreases non-linearly when
repeatedly compressing the image with a lossy codec. This PSNR decrease can be
assessed with a rate-distortion curve. It turns out that recompression of JPEG
AI images can be well detected by this feature under constraints that are
comparable to traditional JPEG forensics methods.
\item \textbf{Quantization cues for distinguishing AI-generated and AI-compressed images.} An important difference between AI-generated and AI-compressed images is the quantization of the AI compressor.
Features that probe for quantization hence enable the discrimination of AI-generated and AI-compressed images.
\end{enumerate}

One notable advantage of the proposed cues is that they are analytic, in the sense that a forensic analyst can relatively easily inspect the details of the calculation (as opposed to a end-to-end neural network with millions of parameters).
Each of these cues is applicable for certain use cases, as shown in our quantitative evaluation.
To the best of our knowledge, none of these cases has been addressed in the literature.
We hope that our work helps to inspire further research on even more powerful forensic methods.

The paper is organized as follows. The Sections 2 to 4 provide contextualization and background. Here, Sec.~2 discusses related work, Sec.~3 briefly introduces JPEG AI compression, and Sec.~4 shows limitations of CNNs for JPEG AI forensics and argues for analytic methods. Sec.~5 introduces the three proposed cues to detect and distinguish JPEG AI images. Sec.~6 reports quantitative evaluations of the three proposed cues. Sec. 7 concludes the work.

\section{Related Work}
\label{sec:relatedwork}
Compression cues have been extensively studied in image forensics. Almost all works study cues for traditional JPEG compression, which started about 20 years ago~\cite{Lukas_03_DJPEG,Popescu_Farid_05}. A minority of works also study other compression formats, such as HEIF~\cite{McKeown_2020}. However, we limit the review of related work to the specific tasks addressed in this work. 

This new JPEG AI compression format presents new challenges for image forensics.
In particular, existing image forensic tools tuned to traditional JPEG images do not generalize well to AI-compressed images~\cite{Berthet_ICIP_AI_Compr}.
Furthermore, AI compression also challenges traditional watermarking algorithms~\cite{AIComprSource}. Cannas~\textit{et al.} show counter-forensic effects in JPEG AI, namely that real images compressed with JPEG AI images are classified as fake by synthetic image detectors, and JPEG AI confuses image splicing localization techniques~\cite{Cannas2024}. Hence, a new set of forensic tools is required for JPEG AI. 

Until now, research in JPEG AI forensics is still scarce.
It has been shown that JPEG AI and state-of-the-art AI codecs leave frequency artifacts in the frequency domain, similarly to synthetic images~\cite{Cannas2024, Bergmann2024}.
Additionally, JPEG AI and AI codecs in general leave visual artifacts in the image, which can also change the semantic content of an image~\cite{Hofer2024, Tsereh2024}.
Furthermore, Kovalev~\textit{et al.} explore the adversarial robustness of JPEG AI~\cite{Kovalev_2024}.
There exist analogies between traditional JPEG forensics and JPEG AI in the case of double compression. Here, a stronger second JPEG AI compression erases traces from the previous compression~\cite{Cannas2024}. In this work, we go one step further and propose analytic features for detecting JPEG AI compression, for distinguishing JPEG AI single- and double-compression, and for distinguishing AI-generated and AI-compressed images. The next paragraphs discuss related work for these three applications.

Several works address the detection of whether an image has previously been compressed with traditional JPEG. For example, the blockwise processing of JPEG leave detectable block artifact grids~\cite{Zhigang_2003}, and also the specific JPEG implementation can cause characteristic artifacts.
For example, rounding operations leave characteristic traces in JPEG compressed images~\cite{Shruti_2020_RoundingJPEG}, and so-called ``JPEG dimples'' arise upon conversion of DCT coefficients from float to integer~\cite{Shruti_2017_Dimples}.
Furthermore, Lorch \textit{et al.} demonstrate that implementation differences in chroma subsampling introduce a high-frequency periodic pattern~\cite{Lorch_2019_Chroma}. In our work, we propose correlations in the color channel as a cue for JPEG AI compression.

Detection of double compression is a widely studied task on JPEG compression~\cite{Benford_JPEG, Lin_2009}.
First works analyzed the DCT coefficients of JPEG images to identify single and double compressed images~\cite{Lukas_03_DJPEG,Popescu_Farid_05}.
More challenging cases of this task are the detection of double compressed images with the same quantization matrix~\cite{Huang_2010, Yang_SameQFJPEG} or on non-aligned JPEG grids~\cite{Bianchi_2012, Barni_NonAligJPEG}.
Here, somewhat related to our specific method are approaches that track the stability of blocks upon recompression~\cite{Yang_SameQFJPEG, Lai2013}, eventhough our proposed cue is notably simpler.
Additionally, methods based on deep neural networks have also been developed to detect double compression~\cite{CATNET, Verma_2023}. Furushita~\textit{et al.} shows the use of coding ghosts to detect double compression for HEIF images~\cite{Furushita_2024}. Cannas~\textit{et al.} presents a first analysis of JPEG AI double compression traces~\cite{Cannas2024}. In our work, we show that the rate-distortion tradeoff of JPEG AI images can be used to detect double-compressed images.

Distinguishing between compressed images and AI-generated images was so far not necessary, because the artifacts from traditional JPEG compression are not very similar to artifacts from synthetic image generators. Instead, JPEG compression usually  leads to robustness issues of deepfake detectors~\cite{Cannas2024}. However, for the case of JPEG AI, Cannas~\textit{et al.} show that JPEG AI images are easily confused as fake by synthetic image detectors~\cite{Cannas2024}. Even retraining the detector on JPEG AI images cannot drastically reduce the false positives. In this work, we propose a quantization cue that is able to distinguish between AI-generated and AI-compressed images. 

\section{JPEG AI Compression}
\label{sec:background}
Early works propose recurrent neural network for AI compression~\cite{Toderici_RNN}.
Contemporary methods are typically based on autoencoders~\cite{Balle_Var_Autoencoder, mbt2018, ms2020}. It is additionally beneficial to incorporate generative adversarial networks (GANs)~\cite{mentzer2020high} or diffusion models to achieve superior image quality at very low bitrates~\cite{Hoogeboom_23}.
Overall, modern methods for AI compression exhibit remarkable compression performance~\cite{JPEGAI_Ascenso}. Hence, the JPEG committee standardized such an end-to-end neural network codec in the new JPEG AI format~\cite{Alshina2024}. 

AI-based image codecs minimize a rate-distortion loss function. The processing chain consists of multiple steps, and we briefly introduce those steps that are relevant for this work.
JPEG AI first converts an input RGB image to the YUV color
space~\cite{Alshina2024}.
The luminance channel Y and the chrominance channels U and V are separated, and the chrominance channels are downsampled
by a factor of 2 (commonly written as 4:2:0 downsampling)~\cite{Alshina2024, Slides_JPEG_AI}.
Luminance and chrominance channels are then compressed with a neural network
$E$. The result of the compression is a latent representation $\mathbf{y}$. Additionally, a latent prediction module processes $\mathbf{y}$ to generate a latent residual and a set of hyperpriors $\mathbf{z}$, which have been first proposed by 
Ball\'{e}~\textit{et al.}\cite{Balle_Var_Autoencoder}.
In this work, we denote an input image as $\mathbf{x}$,
and its encoding with hyperprior as $(\mathbf{y}, \mathbf{z}) = E(\mathbf{x})$. The latent representation is quantized and compressed to a bitstream with an entropy coding modul. 
The decompression of the image inverts the encoding steps, upsamples
chrominance and converts the colors back to RGB~\cite{Alshina2024, Slides_JPEG_AI}. 

JPEG AI defines three version of decoders for decoding the image.
A more complex decoder can produce images with higher quality~\cite{Alshina2024}. 
The standardization of JPEG AI includes specifically the decoders and the latent domain prediction model. The encoder and the hyperprior encoder are not strictly specified and can hence change. Further goals of JPEG AI are to perform computer vision or image processing tasks directly in the latent domain, without the need for explicit decompression~\cite{Alshina2024, Slides_JPEG_AI}. 

\section{Limitations of CNNs for Forensic Analysis}
Deep neural networks are often used as a universal solution for any classification task.
However, deep neural networks often lead to overfitting or biases due to an over-reliance on specific training examples~\cite{Khodabakhsh_2018, Bondi_2020, Corvi_2023}.
Additionally, the use of neural networks can be limited in cases where
decisions of a classification have to be understandable by an analyst or
operator, which can be the case in forensic applications.
While a discriminative neural network can indeed provide high-quality predictions, its ability to explain the influencing factors for the prediction are somewhat limited.

This can become an issue when a forensic analyst has to explain and defend the
result of an analysis. For example, a court case involving synthetically
generated images may call for such a defense. While this issue is difficult in itself, it is further exacerbated by the advent of JPEG AI, which leads to similar traces as synthetic images, and hence both types of images can be easily confused~\cite{Cannas2024}.
Hence, the ability to distinguish AI-compressed from AI-generated images can be expected to gain importance in the near future.
Preventive measures for deep neural networks, such as retraining with JPEG AI images, are not drastically reduce false positive rates~\cite{Cannas2024}.
Therefore, it is important to also develop forensic methods that provide interpretable features to investigate JPEG AI images and to distinguish them from synthetic images.
In this work, we aim to do a first step towards this goal by providing a first set of forensic tools for JPEG AI with interpretable features that are rooted in mechanisms of the JPEG AI algorithm. 

\section{Proposed Cues for JPEG AI}
\label{sec:detection}
We introduce the proposed color correlation cue for detecting JPEG AI compression in Sec.~\ref{subsec:color-correlation-features}, the rate-distortion cue for detecting JPEG AI recompression in Sec.~\ref{subsec:rate_distortion_cues}, and the quantization cue for discriminating AI-generated from AI-compressed images in Sec.~\ref{subsec:quantization_cue}.

\begin{figure*}[tb]
	\centerline{\includegraphics[keepaspectratio,width=\linewidth,height=\linewidth]{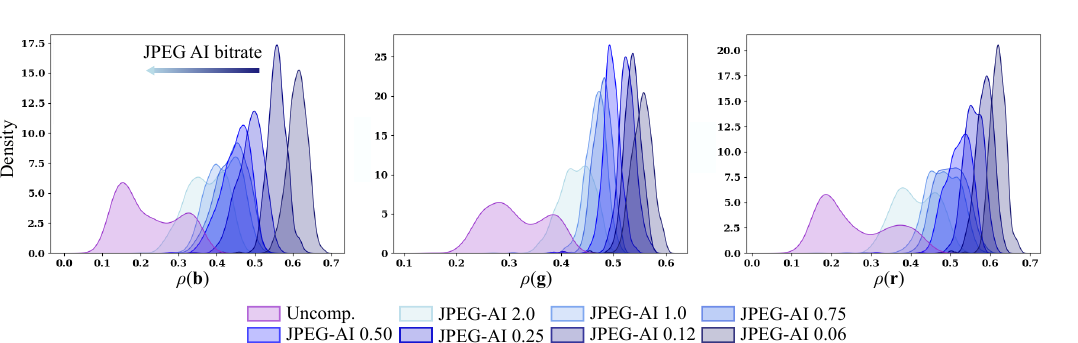}}
	\caption{Color correlations for the three color channels. Uncompressed images (violet) exhibit lower correlations than JPEG AI. The correlations are higher for stronger compression (corresponding to lower bitrates).}
	\label{fig:color_space_feature}
\end{figure*}

\subsection{Color Correlations}
\label{subsec:color-correlation-features}

JPEG AI converts an RGB input image to YUV to separate luminosity and
chromaticity. The chromaticity channels can subsequently be
downsampled for a more compact encoding. Both operations together introduce a slight correlation between the color channels, which can be observed in the high-frequency components of an image. 
We observed earlier that RGB-based
detectors for AI compression are surprisingly resilient to
downsampling~\cite{Bergmann_2023}, which can (in retrospect) be seen as
an indicator of a spectral dependency.

Extracting these correlations is relatively straightforward.
First, we split an image into the three channels red, green, and blue,
and calculate so-called noise-residuals by applying a highpass filter $f$ to
isolate the high-frequency part of each color channel. The noise residuals for
the red, green, and blue channels are denoted as
(linearized) vectors $\mathbf{r}$, $\mathbf{g}$, and $\mathbf{b}$.

To link the noise patterns of the color channels, we pay particular attention
to the differences between these color residuals and the correlations between
these differences. More specifically, let
\begin{align}
\delta(\mathbf{r}, \mathbf{g}) & = |\mathbf{r} - \mathbf{g}| \label{eqn:delta1} \\
\delta(\mathbf{g}, \mathbf{b}) & = |\mathbf{g} - \mathbf{b}| \\
\delta(\mathbf{r}, \mathbf{b}) & = |\mathbf{r} - \mathbf{b}|
\end{align}
denote the differences between color residuals. Then, the normalized
cross-correlation between $\delta(\mathbf{r}, \mathbf{g})$ and $\delta(\mathbf{g}, \mathbf{b})$ 
is defined as
\begin{equation}
\rho(\mathbf{g}) = \frac{\langle \delta(\mathbf{r}, \mathbf{g}), \delta(\mathbf{g}, \mathbf{b})\rangle}{\|\delta(\mathbf{r}, \mathbf{g})\| \| \delta(\mathbf{g}, \mathbf{b})\|}\enspace. \label{eqn:color_correlation}
\end{equation}
Here, by a slight abuse of notation, only the color channel $\mathbf{g}$ appears as argument to $\rho$, but not the other two color channels (whose order is exchangeable due to symmetries in Eqns.~(\ref{eqn:delta1}) to (\ref{eqn:color_correlation})).
The correlations for the two other color channel combinations $\rho(\mathbf{r})$ and $\rho(\mathbf{b})$ are defined analogously.

In all practical experiments, we collect in the vectors $\mathbf{r}$, $\mathbf{g}$, $\mathbf{b}$ only the residuals from a single pixel row of a centrally cropped $512\times 512$ pixels patch of an image. Hence, a single correlation describes a single patch row, and we concatenate the correlations from all rows in a $512$-dim. feature vector.

We illustrate the distinctiveness of these color correlations in a small experiment.
200 uncompressed images are randomly selected from the RAISE
dataset~\cite{RAISE}. From each image, a patch of $512\times 512$ pixels is
cropped. Each such patch is used as-is (``uncompressed'') and with compression 
using JPEG AI at bitrates $[0.06,
0.12, 0.25, 0.50, 1.0, 2.0]$. All experiments in this work are done with the
JPEG AI reference software version 7.0 with all tools enabled and with high
operation point configuration~\cite{jpegai}. From each row of each of these
patches, correlations $\rho(\mathbf{r})$, $\rho(\mathbf{g})$, $\rho(\mathbf{b})$ are calculated. The
distributions of these correlations are shown in 
Fig.~\ref{fig:color_space_feature}. Uncompressed images (violet)
exhibit correlations between $0.1$ and about $0.45$.
JPEG AI images exhibit correlations between $0.3$ and about $0.65$.
The correlations depend on the choice of color channel and on the
compression bitrates, where lower bitrates (stronger compression) lead
to stronger correlations.

\begin{figure*}[tb]
	\centerline{\includegraphics[keepaspectratio,width=\linewidth,height=\linewidth]{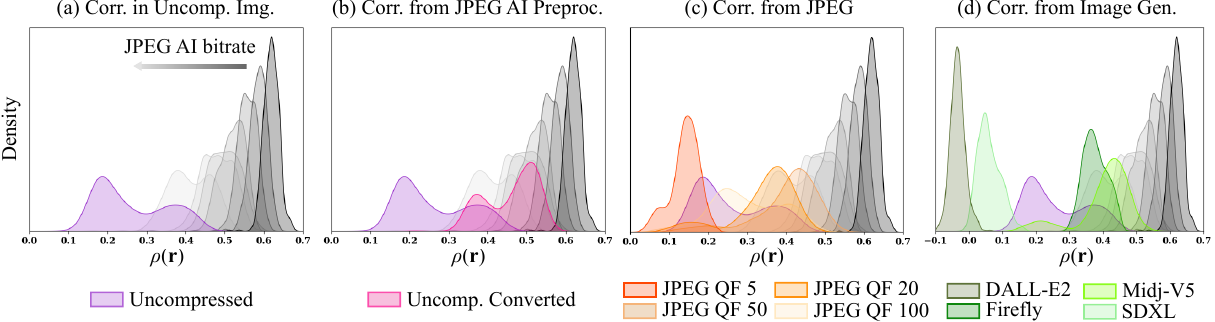}}
	\caption{Distinctiveness of color correlations, shown on the example of $\rho(\mathbf{r})$: (a) reference distributions uncompressed and JPEG AI, as on the right of Fig.~\ref{fig:color_space_feature}. (b) JPEG AI color conversion and 4:2:0 chroma subsampling on uncompressed images (pink) notably increases the correlations over uncompressed images (violet). (c) JPEG compression qualities 20 to 100 slightly increase correlations. (d) The synthetic image generators Midjourney-V5 and Firefly slightly increase correlations.}
	\label{fig:color_space_feature_comparison}
\end{figure*}

Two aspects must be clarified to further characterize the color
correlation cue. First, we show empirical evidence that the correlation
is indeed caused by the color conversion and the downsampling in the
JPEG AI preprocessing. Second, we show how distinctive these
correlations are in comparison to other types of processing, namely
traditional JPEG compression and selected synthetic image generators.
The associated experiments are shown in Fig.~\ref{fig:color_space_feature_comparison} for correlations $\rho(\mathbf{r})$ around the red channel. The other correlations behave analogously.

Fig.~\ref{fig:color_space_feature_comparison}~(a) restates for visual reference the color correlations $\rho(\mathbf{r})$ on uncompressed and JPEG AI-compressed images as previously seen in Fig.~\ref{fig:color_space_feature} (right).

Fig.~\ref{fig:color_space_feature_comparison}~(b) shows that the
observed correlations can indeed be attributed to the preprocessing of
JPEG AI: we apply to each uncompressed patch only the preprocessing of
JPEG AI, i.e., conversion from RGB to YUV, followed by horizontal and
vertical downsampling of the chroma components (i.e., 4:2:0
subsampling). The patches are then upsampled again and converted back
to RGB. The distribution of $\rho(\mathbf{r})$ for these patches (shown
in pink) is notably increased over the uncompressed patches (violet),
and well within the range of JPEG AI correlations. While we note that
stronger JPEG AI compression exhibits even stronger correlations than
preprocessing alone, the preprocessing alone increases the correlations
by at least $0.2$.

Fig.~\ref{fig:color_space_feature_comparison}~(c) examines correlations
induced by traditional JPEG compression. JPEG compression uses a similar preprocessing chain by converting RGB to YCbCr and applying chroma subsampling. The uncompressed patches are compressed with JPEG quality factors $5$, $20$, $50$, and $100$. During compression, we ensure that the same 4:2:0 chroma subsampling is used as for JPEG AI. 
%Figure~\ref{fig:color_space_feature_comparison} (b) shows again the correlations $\rho(\mathbf{r})$ of uncompressed and JPEG AI compressed images, and also of standard JPEG compressed images with quality factors of 5, 20, 50 and 80.
For reasonably high quality factor of $100$, the color correlations slightly increase. With stronger compression of quality factor $50$ the color correlations are almost as much increased as it could be observed for the JPEG AI preprocessing. This is plausible, given the overall similarity of the preprocessing. It is interesting to note that for very strong compressions of quality factors $20$ or even only $5$, the correlations are weakened. We did not further investigate into this effect, but we hypothesize that the large quantization factors and associated large rounding errors act as a decorrelator between the color channels.

Fig.~\ref{fig:color_space_feature_comparison}~(d) examines correlations
in synthetic images. Similar correlations in synthetic images have been reported earlier for some color spaces~\cite{Qiao_2023}. 
Here, we show the correlations for synthetic images from the four
diffusion-based image generators DALL-E2, Midjourney-V5 (Midj-V5),
Firefly, and Stable Diffusion XL (SDXL).
Images from DALL-E2 and SDXL exhibit remarkably low correlations in the range of $[-0.1, +0.15]$.
In contrast, images from Midjourney-V5 and Firefly exhibit correlations that are comparable to traditional JPEG compression and JPEG AI compression at very high bitrates. It may be a coincidence that these generators produce images with more natural color appearance, while DALL-E2 and SDXL tend to generate images that oftentimes look oversaturated. In any case, none of the observed correlations from traditional JPEG or synthetic image generators is as large as the correlations for stronger JPEG AI compression.

Overall, this first study of color channel correlations in JPEG AI images shows that stronger compression exhibits higher similarity between the color channels. The correlation in the color channels can be attributed to the preprocessing and color transformation during JPEG AI compression. Sec.~\ref{sec:detection_color_feature} shows in quantitative experiments that these correlations indeed allow to detect whether an image has been compressed with JPEG AI. 

\begin{figure*}[tb]
	\centerline{\includegraphics[keepaspectratio,width=\linewidth,height=\linewidth]{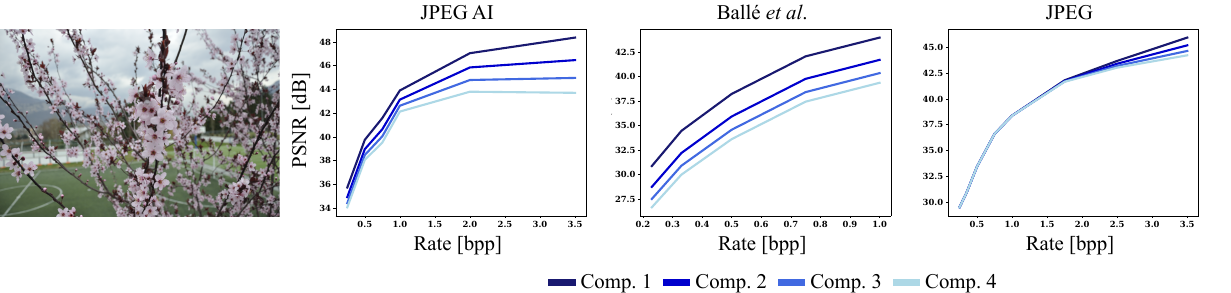}}
	\caption{PSNR curve between original image and compression $k$ per bitrate. We analyze JPEG AI (left), the AI codec by Ball{\'e}~\emph{et al.}~\cite{Balle_Var_Autoencoder} (middle) and JPEG (right). See text for details.}
	\label{fig:original_psnr_vs_rate}
\end{figure*}

\begin{figure*}[tb]
	\centerline{\includegraphics[keepaspectratio,width=\linewidth,height=\linewidth]{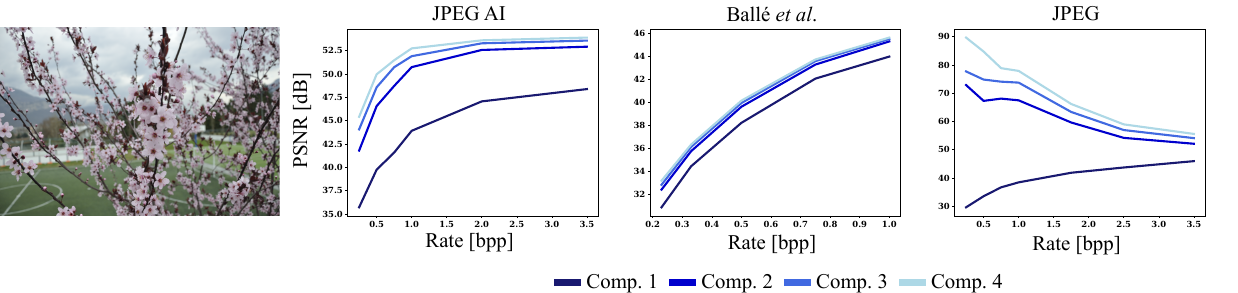}}
	\caption{PSNR between compression $k-1$ and $k$ per bitrate. We analyze JPEG AI (left), the AI codec by Ball{\'e}~\emph{et al.}~\cite{Balle_Var_Autoencoder} (middle) and JPEG (right). See text for details.}
	\label{fig:input_psnr_vs_rate}
\end{figure*}

\begin{figure*}[tb]
	\centerline{\includegraphics[keepaspectratio,width=\linewidth,height=\linewidth]{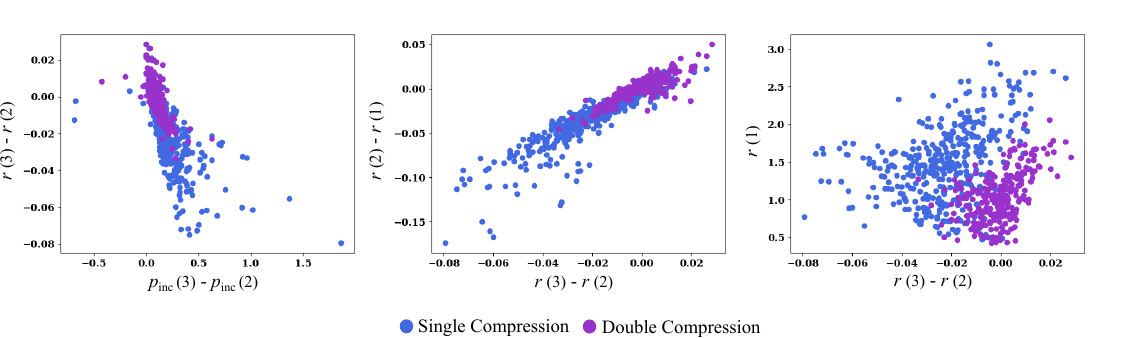}}
	\caption{Dependencies between rate $r(k)$ and PSNR $p_{\mathrm{inc}}(k)$ during $k$ compression runs to differentiate between single and recompressed JPEG AI images. In particular, $r(1)$ and the difference $r(3)-r(2)$ are more discriminative than the others.}
	\label{fig:rate_psnr_feature}
\end{figure*}

\subsection{Rate-Distortion Cue}
\label{subsec:rate_distortion_cues}
Recompression of images with JPEG AI leads to diminishing, but not
negligible changes in the PSNR of the image. This cue can be exploited
for the detection of recompression. In principle, this idea follows the
observation that the amplitudes of frequency artifacts behave similarly
for traditional JPEG and JPEG AI~\cite{Bergmann_2023}. For traditional JPEG, some classical works examine statistics of stable blocks to detect double-compression~\cite{Yang_SameQFJPEG, Lai2013}. These methods are not directly transferable to JPEG AI.
Our proposed approach to the detection of JPEG AI
recompression is hence somewhat different, and from some perspective even simpler: it suffices to track the change of bitrate (bpp) and
PSNR across multiple compressions runs. Every repetition of the
compression leads to smaller changes in the image than the previous
run, depending on the bitrate. The amount of change can be used to identify recompression.

Bitrate and PSNR by itself are not sufficiently descriptive to cover
all possible images. Hence, we show on a single image their dependency.
We will later propose derived features that better generalize across different image content.

The image and a first version of its associated rate-distortion
diagrams are shown in Fig.~\ref{fig:original_psnr_vs_rate}.
The left, middle, and right plots show the diagrams for JPEG AI, a further AI codec by Ball{\'e}~\textit{et
al.}~\cite{Balle_Var_Autoencoder}, denoted as Ball{\'e}~\textit{et al} and traditional JPEG.
The $x$-axis indicates the bitrate of the compressed image.  JPEG AI
allows to directly specify a target bitrate, and for JPEG and
Ball{\'e}~\textit{et al.} we varied the compression quality to achieve
compressions with bitrates between $0.5$ and $3.5$.  An image is
compressed between $1$ (darkblue) to $4$ times (lightblue).
The $y$-axis indicates the PSNR between the uncompressed input image and the image after $k$ compression steps.
For traditional JPEG, repeated compressions barely lead to a change in PSNR, particularly at low bitrates.
Only at a bitrate of $2.0$ or higher, PSNRs are decreased by iterated compression.
For JPEG AI and Ball{\'e}~\emph{et al.}, a loss in PSNR upon repeated compression is noticable across the whole range of investigated bitrates.

It is insightful to study a second type of rate-distortion diagram, which is shown in Fig.~\ref{fig:input_psnr_vs_rate}. 
%%%%
Here, the $y$-axis indicates the
PSNR between two consecutive compressions $k-1$ and $k$. Hence, PSNR is
lowest
between the original image and the 1st compression, because this
compression step changes the image most. Subsequent recompressions have
smaller impact, hence the PSNR between compression steps $k-1$ and $k$
is larger.
This behavior is analogous for all three codecs, and has also been used in
classical works on JPEG recompression forensics~\cite{Yang_SameQFJPEG}.
The PSNR differences of both AI codecs increase with increasing bitrates. In
contrast, JPEG's PSNRs are very large at low bitrates, indicating that
recompression does not change the image much, but flatten for increasing
bitrates somewhere between PSNRs of 45 to 60.
As a sidenote, the differences in PSNR in Fig.~\ref{fig:input_psnr_vs_rate} do
not necessarily lead to a noticable difference in PSNR in
Fig.~\ref{fig:original_psnr_vs_rate} (probably best seen for low JPEG
bitrates), which is rooted in the instable block phenomenon, where individual
JPEG blocks keep changing upon recompression~\cite{Yang_SameQFJPEG, Lai2013}.

Overall, JPEG AI overall exhibits a large PSNR difference between the first and second compression. PSNR differences become smaller at higher bitrates, which is expected.
Also Ball{\'e}~\emph{et al.} exhibit a clear difference between first and second compression, but the differences between second up to fourth compression are relatively small.
Our goal is to define a set of rate-distortion features that robustly describes these observations at different bitrates, in order to distinguish the first and second compression in JPEG AI images.

The feature vector is derived from the rate-distortion curves. First, an input image is three times recompressed with the same compression settings. A $17$-dimensional feature vector is calculated from these recompressed images, consisting of 9 rate features, 4 PSNR features, and 4 differences of rates and PSNRs. More specifically,
\begin{itemize}
\item 9 features are defined by calculating the three rates for each of the three recompressed images: rate $r_{\mathbf{y}}(k)$ of the latent space coefficients $\mathbf{y}$, the rate $r_{\mathbf{z}}(k)$ of the hyperpriors $\mathbf{z}$, and rate $r(k)$ for the whole image calculated as $r(k) = (r_{\mathbf{y}}(k) + r_{\mathbf{z}}(k)) / (h\times w)$, where $h$ and $w$ denote image height and width, and $k$ indicates everywhere the $k$-th recompression.
\item 4 features are obtained from PSNRs. Let $p_{\mathrm{inp}}(k)$ denote the PSNR between the input image and the $k$-th recompression (as in Fig.~\ref{fig:original_psnr_vs_rate}), and let $p_{\mathrm{inc}}(k)$ denote the incremental PSNR between the $k-1$-th and the $k$-th recompression (as in Fig.~\ref{fig:input_psnr_vs_rate}). Then, the features are $p_{\mathrm{inp}}(2)$, $p_{\mathrm{inp}}(3)$, $p_{\mathrm{inc}}(2)$, and $p_{\mathrm{inc}}(3)$.
\item 4 features from the differences $(r(3)-r(2))$, $(r(2)-r(1))$, $(p_{\mathrm{inp}}(3) - p_{\mathrm{inp}}(2))$, and $(p_{\mathrm{inc}}(3)-p_{\mathrm{inc}}(2))$.
\end{itemize}
In our experiments in Sec.~\ref{sec:detection_recompression}, we show that these features even generalize across AI codecs, i.e., we extract these features from the AI codec by Ball{\'e}~\emph{et al.}~\cite{Balle_Var_Autoencoder}.

Figure~\ref{fig:rate_psnr_feature} shows the dependencies between selected features during the compression runs. We observe that the calculated PSNR values and the rate of the three compression runs have a potential to differentiate between single and recompressed JPEG AI images.
Notably, $r(1)$ and the difference $(r(3) - r(2))$ are more discriminative than $(r(2) - r(1))$ and $(p_{\mathrm{inc}}(3) - p_{\mathrm{inc}}(2))$.

\begin{figure}[tb]
	\centerline{\includegraphics[keepaspectratio,width=\linewidth,height=\linewidth]{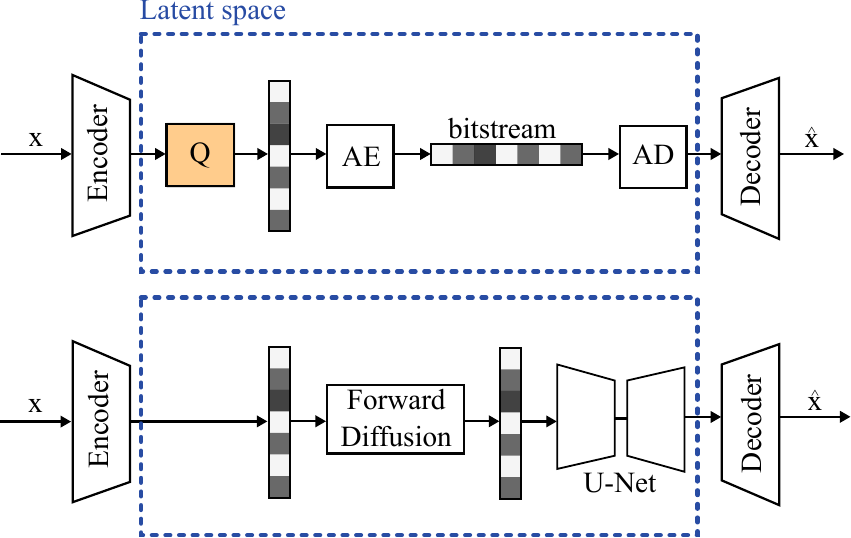}}
	\caption{Architecture comparison of AI codecs (top) and image generators (here latent diffusion, bottom). Compared to image generators a quantization (Q) is applied in the latent space of AI codecs.}
	\label{fig:arch_comparison}
\end{figure}

\subsection{Quantization Cue}
\label{subsec:quantization_cue}
JPEG AI and synthetic images leave similar artifacts in the frequency domain, which can cause misclassifications in synthetic image detectors~\cite{Cannas2024}. 
JPEG AI images can still cause a high false positive rate, even if a synthetic image detector has been retrained with such images.
Thus, it is important to develop methods that can differentiate between AI-compressed and AI-generated images.  

\begin{figure}[tb]
	\centerline{\includegraphics[keepaspectratio,width=\linewidth,height=\linewidth]{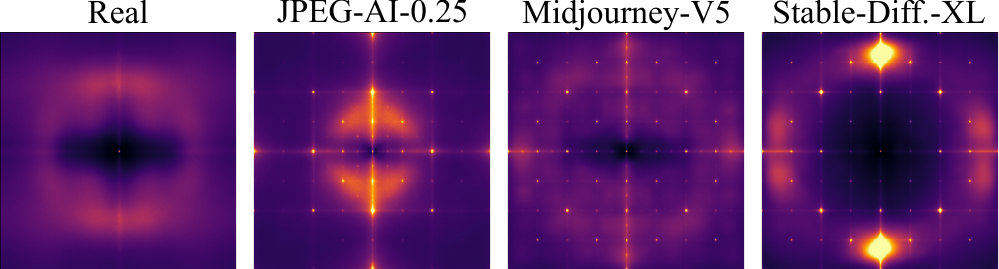}}
	\caption{JPEG AI images and synthetic images show similar artifacts in the frequency domain. We show the average Fourier spectra of real, JPEG AI and diffusion-based image generators (Midjourney-V5, Stable Diffusion XL). Best viewed in the digital version.}
	\label{fig:spectrum}
\end{figure}

One reason for the confusion of AI-compressed and AI-generated images may lie in the fact that their neural network architectures share some similarities.
Fig.~\ref{fig:arch_comparison} (top) coarsely illustrates an architecture of an AI codec, Fig.~\ref{fig:arch_comparison} (bottom) illustrates an architecture of a latent diffusion model. Both architectures perform their core operations in the latent space of an encoder-decoder component. The internal processing differs, and (among further details) consists of a quantizer Q and an arithmetic encoder and decoder for the AI codec, and a forward diffusion process and an U-Net for the image generator.
The decoder, which occurs in both architectures, be responsible for upsampling artifacts that are similar in AI-compressed and AI-generated images. These artifacts have extensively been studied in the literature on synthetic image detection~\cite{Corvi_2022_DM, frank20a_GAN}, and they are illustrated in Fig.~\ref{fig:spectrum}. Here, the Fourier spectra of a real image, an AI-compressed image with bitrate $0.25$, a Midjourney-V5, and a Stable Diffusion XL are shown.
All images except the real image exhibit a regular grid pattern, which can be
attributed to upsampling~\cite{frank20a_GAN}. These similarities in the
patterns may be one reason that AI-compressed and AI-generated images are
easily confused by forensic detectors.

However, both types of images can be distinguished by the fact that only AI codecs include a quantizer Q
(orange box in Fig.~\ref{fig:arch_comparison}). The quantizer rounds the
coefficients of the latent representation of an image to the nearest multiple
of the associated quantization factors. It is a typical component for lossy
image compressors, to optimize the coefficients for arithmetic encoding. 
Hence, a detector for quantization artifacts can also distinguish between AI-compressed and AI-generated images.
As such, quantization features can possibly be combined with any existing synthetic image detector, for example to adjust the confidence of a detector depending on whether the image is compressed with JPEG AI or not.

We propose an indirect detector for quantization, following (very loosely) 
the observation by He \textit{et al.} that the latent representations of
natural images are more robust to added noise in pixel domain than synthetic
images~\cite{He_2024}. In our case, we propose to directly probe for the sensitivity 
to rounding operations in the latent representation of an AI codec. This
follows the assumption that if an images has already been subject to
quantization, then their latent coefficients will exhibit a distinguishable
distribution. 

To extract the quantization features, $256 \times 256$ pixels are cropped
from the center of an input image to ensure that the dimension of the latent
space $\mathbf{y}$ is the same for all inputs. The latent space consists of
$(i, j)$ spatial coefficients, and each coefficient consists of $C$ channels.
Let $\mathbf{y}_{c}$ denote a slice of coefficients for channel $c$. The $c$-th feature is the correlation
\begin{equation}
\phi(\mathbf{y}_{c}) = \frac{\langle \mathbf{y}_{c}, \lfloor \mathbf{y}_{c} \rceil \rangle}{\|\mathbf{y}_{c}\| \cdot \| \lfloor \mathbf{y}_{c} \rceil \|}
\label{eqn:rounding}
\end{equation}
over the spatial coordinates $(i, j)$, where
$\lfloor \mathbf{y}_{c}\rceil$ denotes $\mathbf{y}_{c}$ after element-wise rounding to the nearest integer.

These correlations capture two types of properties.
First, they are sensitive to zero entries (as a zero entry in one vector cancels the entry in the other vector), and second they tend to reward similar step sizes.
Zero entries are arguably most relevant for detecting lossy compression (aiming to maximize sequences of zeros for the arithmetic encoding), but also off-zero entries play an important role.
To show this, we also compute a correlation on binarized vectors, i.e., we truncate all non-zero entries to either $+1$ or $-1$. Our ablation in the experiments in Sec.~\ref{subsec:exp_quantization} shows that this truncated version alone, which is only sensitive to zeros, works surprisingly well but nevertheless notably worse than the full vector.
These experiments also show that the concept hold also across AI codecs,
i.e., that JPEG AI can be distinguished from synthetic images even if the
features are extracted on channel correlations from the variable bitrate version of the encoder by
Ball{\'e}~\emph{et al.}~\cite{Kamisli2024}.

The role of coefficients that are quantized to zero is further illustrated in Table~\ref{tab:quantize_avg_similarity}. Here, we show the average correlation between $\mathbf{y}_c$ and $\lfloor \mathbf{y}_c \rceil$ for JPEG AI images compressed with various bitrates and synthetic images.
For synthetic images we use also the average over Stable Diffusion (XL and V1.4), Glide, Midjourney-V5, Firefly and DALL-E2 images.
For stronger JPEG AI compression, the correlation between $\mathbf{y}_c$ and $\lfloor \mathbf{y}_c \rceil$ decreases. While this is counter-intuitive at first glance, it is a by-product of the quantization.
Stronger compression (i.e., lower bpp) leads to more latent space coefficients close to 0, which are rounded to $0$ in $\lfloor \mathbf{y}_c\rceil$. These are not isolated incidents, but for lower bpps, whole channels $c$ are set to $0$, which induces a correlation for that channel of $0$.
In contrast, synthetic images are much less affected by this effect. 

\begin{table}[!t]
	\renewcommand{\arraystretch}{1.0}
	\caption{Mean correlation of quantization features for different JPEG AI bitrates (bpps: 0.06 to 0.75) and synthetic images (see text for details). Stronger compression causes lower mean correlation, which seems counter-intuitive, but stems from whole channels being quantized to $0$.}
	\label{tab:quantize_avg_similarity}
	\centering
	\begin{tabular}{p{2.0cm}|p{0.5cm}p{0.5cm}p{0.5cm}p{0.5cm}p{0.5cm}|p{0.8cm}}
		\toprule
		 & 0.06 & 0.12 & 0.25 & 0.50 & 0.75 & Synthetic  \\
		\midrule
		Mean similarity & 0.486 & 0.576 & 0.690 & 0.770 & 0.812 & 0.835 \\
		\bottomrule
	\end{tabular}
\end{table}

\section{Evaluation}
This Section provides quantitative evaluations for the three proposed cues for JPEG AI forensics.
Sec.~\ref{sec:detection_color_feature} evaluates the detection JPEG AI compression with color correlations.
Sec.~\ref{sec:detection_recompression} evaluates the detection of double compression in JPEG AI images with rate-distortion features).
Sec.~\ref{subsec:exp_quantization} evaluates the discrimination of JPEG AI and synthetic images with quantization features.

\subsection{Detection of JPEG AI Compression}
\label{sec:detection_color_feature}
This experiment quantitatively shows the suitability of color correlations for the detection of JPEG AI compression. The color correlations are fed to a simple random forest. As a baseline to understand the overall difficulty of the task, we also train a ResNet50~\cite{He_2016_ResNet} in the pixel domain to distinguish uncompressed and JPEG AI compressed images.

\subsubsection{Experimental Setup and Datasets}
The performance of the color correlations is evaluated on the RAISE dataset, which consists of 1000 uncompressed, high-quality images~\cite{RAISE}.
For training the random forest and the ResNet50, the dataset is split into 700 images for training, and 200 images for testing. We additionally use 100 images for validation for the ResNet50.
For the training, a copy of each uncompressed training image is compressed with
JPEG AI at a bitrate of $0.5$. All images are centrally cropped to $512\times
512$ pixels patches to ensure identical dimensions everywhere, and to simplify the
evaluation.

The color correlations process each patch line-by-line, which leads to a
$512$-dimensional feature vector per correlation $\rho(\mathbf{r})$,
$\rho(\mathbf{g})$, $\rho(\mathbf{b})$. These
feature vectors are fed to a random forest with 500 trees.
As a baseline comparison, we also train a ResNet50 without the color
correlations directly on the $512\times 512$ pixels patches.
The ResNet50 training runs for 5 epochs using an Adam optimizer with a learning
rate of 0.001 and a batch size of 8.

The 200 images in the test set are used as uncompressed images, and as JPEG AI compressed copies with
the bitrates $0.06$, $0.12$, $0.25$, $0.50$, $0.75$, $1.0$, and $2.0$. Testing images are also
centrally cropped to $512\times512$ pixels.

\begin{table}[!t]
	\renewcommand{\arraystretch}{1.0}
	\caption{Accuracy of color correlation features to detect JPEG AI images. The proposed correlations perform better at lower bitrates (bpp), and generalize better to unseen bpp.}
	\label{tab:results_color_features}
	\centering
	\begin{tabular}{p{0.9cm}p{0.6cm}p{0.6cm}p{0.6cm}p{0.6cm}p{0.6cm}p{0.6cm}p{0.6cm}}
		\toprule
		\textbf{bpp} & \textbf{0.06} & \textbf{0.12} & \textbf{0.25} & \textbf{0.50} & \textbf{0.75} & \textbf{1.0} & \textbf{2.0} \\
		\midrule
		ResNet50 & 0.834 & \textbf{0.859} & \textbf{0.875} & \textbf{0.855} & \textbf{0.801} & 0.745 & 0.673  \\
		$RF \rho(\mathbf{r})$ & 0.822 & 0.797 & 0.774 & 0.744 & 0.738 & 0.726 & 0.667  \\
		$RF \rho(\mathbf{g})$ & 0.829 & 0.794 & 0.775 & 0.743 & 0.716 & 0.702 & 0.658  \\
		$RF \rho(\mathbf{b})$ & \textbf{0.856} & 0.829 & 0.811 & 0.785 & 0.761 & \textbf{0.746} & \textbf{0.710}   \\
		\bottomrule
	\end{tabular}
\end{table}

\subsubsection{Results for Color Correlation Features}
Table~\ref{tab:results_color_features} shows the results of this experiment.
Among the three variants of the color correlations, the correlation
$\rho(\mathbf{b})$ centered around the blue channel performs best. Across all
tested bitrates, it achieves detection rates range between $71.0\%$ and $85.6\%$,
which appears reasonable given that the training was only performed on bitrate
$0.5$. The lowest performances occur at the highest bitrates, which is also
plausible considering the reduced impact of compression at higher bitrates (cf.\ also Fig.~\ref{fig:color_space_feature}).
The baseline ResNet50 classifier performs better for bitrates that are
close to the training bitrate of $0.5$ and worse for bitrates that are far away
from the training bitrate. This is plausible, given that neural networks
have a tendency to learn specific tasks well, but they oftentimes generalize
worse to unseen data. Besides generalization, a benefit of the color
correlations over ResNet50 is the simplicity and interpretability of the
feature, which can open opportunities for a forensic analyst to further
investigate the result. Nevertheless, for a practical application, one would
desire performances beyond $90\%$ accuracy, which underlines the need to
perform further research on this subject in the future.

We close this subsection with a small experiment on the possible
confusion of uncompressed images with JPEG AI preprocessing, as
examined earlier in Fig.~\ref{fig:color_space_feature_comparison}~(b).
When classifying such preprocessed images with the pre-trained random
forest, its mean accuracy drops by $12\%$ from $78.0\%$ to $66.0\%$.
Including these images into the training restores the mean accuracy,
which on one hand indicates that preprocessing is an important contributor factor, but on the other hand it is not the only contributing factor for correlating the color channels.

\subsection{Detection of JPEG AI Recompression}
\label{sec:detection_recompression}
This experiment shows the performance of the rate-distortion features from Sec.~\ref{subsec:rate_distortion_cues} for the purpose of distinguishing single- versus double-compression in JPEG AI images. For this task, we fed the rate-distortion features to a random forest and as a baseline we additionally train a ResNet50~\cite{He_2016_ResNet} in the pixel domain.

\begin{table}[!t]
	\renewcommand{\arraystretch}{1.0}
	\caption{Accuracy of a random forest (RF) trained on rate-distortion features to detect recompressed JPEG AI images. Rate-distortion features generalize better than the pixel-based ResNet50, particularly when $b_0 < b_1$, i.e., $b_0 < 0.75$.}
	\label{tab:results_recompress}
	\centering
	\begin{tabular}{p{0.5cm}p{0.8cm}p{0.8cm}p{0.8cm}p{0.8cm}}
		\toprule
		& \multicolumn{2}{c}{\textbf{ResNet50} (Pixel)}  & \multicolumn{2}{c}{\textbf{RF} (Rate-Dist.)} \\
		\midrule
		$b_0$  & Single & Recomp. & Single & Recomp.\\
		\midrule
		2.00 &  0.797 &  0.476  &  0.913 &  0.511 \\
		1.00 &  0.764 &  0.475  &  0.912 &  0.518 \\
		0.75 &  0.726 &  0.532  &  0.879 &  0.548 \\
		0.50 &  0.644 &  0.616  &  0.777 &  0.725 \\
		0.25 &  0.519 &  0.726  &  0.526 &  0.879 \\
		0.12 &  0.457 &  0.778  &  0.465 &  0.920 \\
		\bottomrule
	\end{tabular}
\end{table}

\subsubsection{Experimental Setup and Datasets}
For the training of the random forest and the ResNet50 we use 450 images for training and 200 images for testing from the RAISE dataset~\cite{RAISE}. For the ResNet50, we additionally use 100 images for validation.
A random forest classifier is trained on the proposed rate-distortion features with 
every training image is once single-compressed images with bitrate $b_0 = 0.75$, and once double-compressed images with bitrates $b_0 = 0.25$ followed by $b_1 = 0.75$.
The rate-distortion features are extracted with the AI codec by Ball{\'e}~\textit{et al.}~\cite{Balle_Var_Autoencoder} in order to underline that the proposed feature set is not specific to JPEG AI. To extract the features, we operate the AI codec by Ball{\'e}~\textit{et al.} at quality level 8. 
We train a random forest with 500 trees on the extracted rate-distortion features with the
same standard settings as above, and a ResNet50 directly
on the JPEG AI-compressed images in the pixel domain on the default $224 \times
224$ pixels patches. The training again runs for 5 epochs using an Adam
optimizer with a learning rate of 0.001 and a batch size of 8.

The 200 testing images are prepared as follows. Every image is compressed at the bitrates
$b_0 \in \{0.12, 0.25, 0.50, 0.75, 1.0, 2.0\}$. We store these images as single-compressed, and then recompress copies of each of these images with a second bitrate $b_1 = 0.75$.
Hence, most of the evaluated bitrates are unseen during training, to show the generalization capability of the proposed features.
In particular, these settings cover the cases of $b_0 < b_1$,  $b_0 > b_1$, and
$b_0 = b_1$, which are commonly distinguished  in traditional JPEG
double-compression analysis. To balance the number of positives and negatives, we group single-compressed and double-compressed images in a particular way. For evaluating the detection of single-compression, we compare single-compressed images with arbitrary $b_0$ versus double-compressed images with fixed $b_0 = 0.25$ and $b_1 =0.75$. Conversely, for evaluating the detection of double-compression, we compare single-compressed images with fixed $b_0=0.75$ versus double-compressed images with arbitrary $b_0$ and fixed $b_1 =0.75$.

\subsubsection{Results for Rate-Distortion Features}
The results are shown in Table~\ref{tab:results_recompress}.
In general, detecting recompression with bitrates $b_0 \ge 0.75$ is challenging for both classifiers, where also the better performing rate-distortion features are only marginally better than guessing. This tendency is in line with similar results from traditional JPEG forensics, where a second, stronger compression usually erases the traces of a weaker primary compression. Both classifiers tend to detect single-compressed images at these bitrates correctly, whereby the proposed rate-distortion features perform better than the pixel-based ResNet50. In case when $b_0 < 0.75$, double-compression can be well detected, but single-compression (due to the out-of-distribution characteristics of the training data) is increasingly challenged to correctly detect these cases.

\subsection{JPEG AI vs. Synthetic Images}\label{subsec:exp_quantization}
This experiment shows the effectiveness of the quantization features to distinguish between JPEG AI images and synthetic images.
 We fed the quantization features to a simple random forest and as a baseline we again train a ResNet50~\cite{He_2016_ResNet} in the pixel domain.
 
\subsubsection{Experimental Setup and Datasets}
\label{subsec_quantize_experiment_setup_datasets}
The training is performed on a subset of the Synthbuster
dataset~\cite{Synthbuster}, which includes uncompressed images from the RAISE
database and synthetic images from several diffusion generators.  For the
training of the random forest and the ResNet50, we use 700 real images from
RAISE, and 700 synthetic images from Stable Diffusion XL. For the ResNet50, we
additionally use 100 images per class for the validation set. The uncompressed
images are compressed with JPEG AI at a bitrate of $0.25$.
The quantization features are computed on a central patch of $256 \times 256$ per image to ensure identical dimensions in the latent space of the AI codec. For feature extraction, we use the variable bitrate version of the encoder by Ball{\'e}~\textit{et al.} at quality level 4~\cite{Kamisli2024} to show the transferability of the features across encoders, which has $C=320$ channels per spatial coefficient $(i,j)$.
We train the random forest with 500 trees on two types of correlations, namely on the correlation of the quantized vectors in Eqn.~(\ref{eqn:rounding}), and on the 0-1 truncated version of these vectors (cf.\ Sec.~\ref{subsec:quantization_cue}). As a baseline, we train a ResNet50 on $224 \times 224$ patches for 5 epochs using a Adam optimizer with a learning rate of 0.001 and a batch size of 8.

The testing is also performed on images from the Synthbuster dataset~\cite{Synthbuster}. We use 200 other images from the RAISE dataset as real images, and 200 images from each of the available synthetic data sources, which are DALL-E2, Firefly, Glide, Midjourney-V5, Stable Diffusion 1.4, and Stable Diffusion XL. 
We additionally collect from the dataset by Corvi~\textit{et al.} 200 images
per GAN-based synthetic image generator, namely BigGAN, StyleGAN2, StyleGAN3
and ProGAN~\cite{Corvi_2022_DM}. In all experiments, all real images that are used in testing are compressed with 
lower JPEG AI bitrates of $0.06$, $0.12$, $0.25$, $0.50$ and $0.75$, which provoke more false positives, as reported by Cannas~\textit{et al.}~\cite{Cannas2024}.

We report the detection accuracies of either specific compression strengths or specific image generators by setting up a two-class classification goal. A target compression strength is classified versus Stable Diffusion XL images, and a target synthetic image generator is classified versus JPEG AI-compressed RAISE images at bitrate $0.25$.

\begin{table}[!t]
	\renewcommand{\arraystretch}{1.0}
	\caption{Accuracy of a random forest (RF) on quantization feature (Trunc.: truncation, full: full integer vector) to differentiate between JPEG AI and synthetic images. The results of the simple classifiers show a higher generalization than a ResNet trained on pixels.}
	\label{tab:quantize_results}
	\centering
	\begin{tabular}{p{1.3cm}p{1.5cm}p{1.2cm}p{1.0cm}p{0.9cm}}
		\toprule
		\textbf{Type} &  &  \textbf{ResNet50} (Pixel)  & \textbf{RF} (Trunc.) & \textbf{RF} (full) \\
		\midrule
		\multirow{4}{*}{\small \textbf{JPEG AI}} & \small bpp \small 0.06 & \small 0.886  & \small 0.905  & \small \textbf{0.985} \\
		& \small bpp \small 0.12 & \small 0.886  & \small 0.896  & \small \textbf{0.982} \\
		& \small bpp \small 0.25 & \small 0.879   & \small 0.849 & \small \textbf{0.977} \\
		& \small bpp \small 0.50 & \small 0.850   & \small 0.782  & \small \textbf{0.959} \\
		& \small bpp \small 0.75 & \small 0.827  & \small 0.731 & \small \textbf{0.917} \\
		\midrule						 
		\multirow{5}{*}{\small \textbf{DM}} & \small Midj.-V5 & \small 0.850 & \small 0.832  & \small \textbf{0.945}   \\
		& \small DALL-E2  & \small 0.871  & \small 0.789  & \small \textbf{0.950}\\
		& \small SD.-XL & \small 0.879  & \small 0.849  & \small \textbf{0.977} \\
		& \small SD.-1.4 & \small 0.869  & \small 0.936  & \small \textbf{0.980} \\
		& \small Glide & \small \textbf{0.661}  & \small 0.489  & \small 0.538 \\
		& \small Firefly & \small  0.708   & \small 0.572  & \small \textbf{0.839} \\
		\midrule
		\multirow{4}{*}{\small \textbf{GAN}} & \small BigGAN & \small 0.693 & \small 0.765  & \small \textbf{0.767} \\ 
		& \small StyleGAN2 & \small 0.748  & \small 0.775  & \small \textbf{0.853}  \\
		& \small StyleGAN3 & \small 0.798  & \small 0.839  & \small \textbf{0.859} \\
		& \small ProGAN & \small 0.723  & \small 0.854  & \small \textbf{0.980} \\
		
		\bottomrule
	\end{tabular}
\end{table}

\begin{figure}[tb]
	\centerline{\includegraphics[keepaspectratio,width=\linewidth,height=\linewidth]{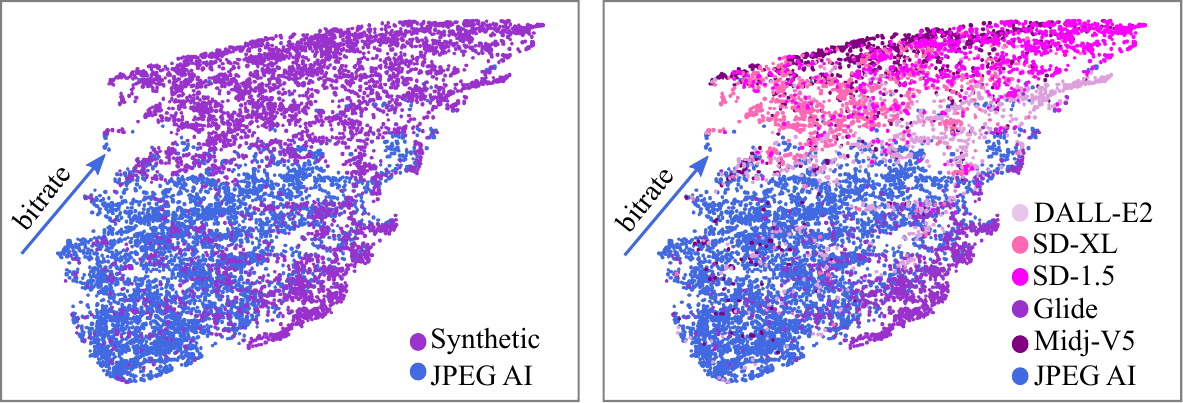}}
	\caption{2D UMAP visualization of JPEG AI and synthetic images with quantization features. The left feature space shows that lower JPEG AI bitrates are more separable to synthetic images. The right feature space reveals the separability on different image generators.}
	\label{fig:umap_quantize}
\end{figure}

\subsubsection{Results for Quantization Features}
Table~\ref{tab:quantize_results} shows the results for the detection of JPEG AI images, diffusion-based and GAN-based synthetic images.
The baseline ResNet50 performs best for Glide images with an accuracy of 66.1\%, but otherwise it generalizes worse than the quantization features.
The quantization features exhibit a good generalization across compression strengths and generator types. The importance of the coefficients that are quantized to zero are shown in the very respectable performance of the truncated features (Trunc.), which in many cases perform comparable to the ResNet50 classifier.
However, quantization features that use the untruncated, full integer vector (full) still perform notably better. 
These results confirm that the amount of zeros after quantization is an important cue for differentiating AI-compressed and AI-generated images. Nevertheless, it also shows that also other factors contribute. 
The accuracy of the full vector for detecting JPEG AI is for all bitrates over 91.0\%, and stronger compression leads to higher accuracies. Diffusion-based images are very reliably distinguished from JPEG AI images with accuracies beyond $94\%$ for four out of six generators. GAN-based images perform worse, which can also be attributed to the fact that it is generally difficult to generalize from diffusion-based images to GAN-based images and vice versa~\cite{Corvi_2022_DM}.

Figure~\ref{fig:umap_quantize} shows a UMAP visualization of the feature space of the full integer vector quantization features.
The left plot shows that JPEG AI and synthetic images are separable, and that in particular lower bitrates differentiate JPEG AI images further from synthetic images.
The right plot further differentiates this feature space into the different image generators. 
In most cases, the feature points of the image generators cluster above the JPEG AI images.
Interestingly, the Glide images are mapped to the right side of the JPEG AI features. This outlier behavior of Glide is in line with the quantitative experiments, where Glide images achieve by far the lowest detection accuracies.

\begin{table}[!t]
	\renewcommand{\arraystretch}{1.0}
	\caption{Robustness of the quantization features to postprocessing attacks JPEG (JPG) and resizing (RS).}
	\label{tab:robustness_results}
	\centering
	\begin{tabular}{p{1.2cm}p{0.8cm}p{0.8cm}p{0.8cm}p{0.7cm}p{0.7cm}p{0.7cm}}
		\toprule
		 JPEG AI & JPG 70 & JPG 80 & JPG 90 & RS 70 &  RS 80 & RS 90  \\
		\midrule
		bpp 0.25 & 0.931 & 0.955 & 0.985 & 0.908 & 0.972 & 0.980  \\
		bpp 0.50 & 0.880 & 0.885 & 0.978 & 0.732 & 0.908 & 0.962 \\
		bpp 0.75 & 0.805 & 0.791 & 0.840 & 0.631 & 0.775 & 0.928\\
		\midrule
		Midj.-V5 & 0.817 & 0.841 & 0.873 & 0.690 & 0.861 & 0.915  \\
		Firefly  & 0.673 & 0.738 & 0.865 & 0.615 & 0.648 & 0.675  \\
		DALL-E2  & 0.720 & 0.813 & 0.883 & 0.715 & 0.873 & 0.921 \\			 
		\bottomrule
	\end{tabular}
\end{table}

Additionally, we evaluate the robustness of the quantization features to postprocessing, namely, JPEG compression and resizing.
We only add postprocessing to a subset of the test data, namely on JPEG AI
bitrates $0.25$, $0.50$ and $0.75$ and generator data from Midjourney-V5,
Firefly, and DALL-E2. Copies of these images are compressed with traditional
JPEG at quality factors 70, 80, and 90 and downsampled to 70\%, 80\%, and 90\%
of the original size. Table~\ref{tab:robustness_results} shows the performance
of the full quantization features on these images. Not surprisingly, we observe
that with stronger JPEG compression and resizing the accuracy decreases.
Nevertheless, the performance declines gently, such that one may assume that
further investigations can develop more advanced features for distinguishing
AI-compressed from AI-generated images.

\section{Conclusion}
\label{sec:conclusion}
In this work, we provide a first step towards forensic tools for JPEG AI.
First, we show that JPEG AI characteristically correlates the color channels, which can be used to detect when an image has been compressed in the JPEG AI format.
Second, we show that rate-distortion curves expose differences in single-compressed and double-compressed JPEG AI images. Third, we show that the quantization in JPEG AI enables the distinction between AI-compressed and AI-generated images. The proposed cues are interpretable, and we hope that they on one hand are useful for forensic analysts, and on the other hand inspire further research in compression forensics of JPEG AI.

\bibliographystyle{IEEEtran}
\bibliography{IEEEabrv, references}

\EOD

\end{document}